\documentclass{llncs}
\usepackage{makeidx}  
\usepackage{graphicx}
\usepackage{subcaption}
\usepackage{cite}
\usepackage{authblk}
\begin{document}
\frontmatter          
\pagestyle{headings}  
\addtocmark{A 3D-CNN for Functional Connectivity based classification} 
\title{3D Convolutional Neural Networks for Classification of Functional Connectomes}
\titlerunning{CNNs for Connectome Data}  
%
\author{Meenakshi Khosla \inst{1} \and Keith Jamison \inst{2,3} \and Amy Kuceyeski \inst{2,3} \and Mert R. Sabuncu \inst{1,4}}


%
%
%
\institute{1. School of Electrical \& Computer Engineering, Cornell University\\
2. Radiology, Weill Cornell Medical College \\
3. Brain and Mind Research Institute, Weill Cornell Medical College \\
4.  Nancy E. \& Peter C. Meinig School of Biomedical Engineering, Cornell University}

\maketitle              

\begin{abstract}
 
Resting-state functional MRI (rs-fMRI) scans hold the potential to serve as a
diagnostic or prognostic tool for a wide variety of conditions, such as autism, Alzheimer's disease, and stroke. 
While a growing number of studies have demonstrated the promise of machine learning algorithms for rs-fMRI based clinical or behavioral prediction, most prior models have been limited in their capacity to exploit the richness of the data.
For example, classification techniques applied to rs-fMRI  often rely on region-based summary statistics and/or linear models.
In this work, we propose a novel volumetric Convolutional Neural Network (CNN) framework that takes advantage of the full-resolution 3D spatial structure of rs-fMRI data and fits non-linear predictive models. 
We showcase our approach on a challenging large-scale dataset (ABIDE, with $N>2,000$) and report state-of-the-art accuracy results on rs-fMRI-based discrimination of autism patients and healthy controls.
\end{abstract}
\section{Introduction}
%

The connectome, which can be captured via neuroimaging techniques such as diffusion and resting-state functional MRI, is a research area of intense focus, as it has delivered and continues to promise novel neuroscientific insights and clinical tools.
In recent years, machine learning algorithms have been increasingly applied to connectome data~\cite{Plitt2015, Mennes2012, varoquaux10}.
These models often employ hand-engineered features such as pairwise correlations between regions of interest (ROIs) and network topological measures of clustering, small-worldness, integration, or segregation~\cite{BrownH16,Kaiser2011}.
Furthermore, a vast majority of these models collapse the data into a feature vector for use in standard classification algorithms. 
Vectorization, however, discards the spatial structure of the connectome, which is an important source of predictive information~\cite{kong2018spatial}. 
Finally, many machine learning techniques used with connectome data rely on linear or ``shallow'' models, which are limited in their capacity to capture relationships between connectomic features and clinical/behavioral variables.
    
%

In related work, deep neural networks exploiting the topological properties of brain networks have been recently explored. 
For example, the BrainNetCNN architecture of~\cite{Kawahara2017} extends convolutional neural networks (CNNs) to handle graph-structured data. 
While CNNs are motivated via the translation-invariance property of image-based classification problems and thus have achieved tremendous success, the neuroscientific basis of the invariance property exploited by BrainNetCNN remains elusive.
Furthermore, this approach works directly with an adjacency matrix derived from the connectome data, while disregarding spatial information. 
Graph convolution networks~\cite{pmlr-v48-niepert16}, while increasingly popular, also seem sub-optimal to use with connectome data, since they rely on a common graph and the variation of interest is in the node properties.
In the connectome, however, the main variation is the adjacency matrix, i.e., edge properties.  




Our core contribution is an easy-to-implement 3D CNN framework for connectome-based classification. Our key insight is to use the connectivity ``fingerprint'', or functional coupling of each voxel to distinct target ROIs, as input features for a traditional volumetric CNN, represented as a multi-channel image volume. 
This allows us to characterize connectivity at a much finer scale than previously used with machine learning techniques, and without losing the spatial relationship between voxels.
We are agnostic to the exact definition of target ROIs, yet as we demonstrate empirically this choice can impact final accuracy. 
In our experiments, we present an ensemble learning strategy that averages models obtained with different ROI definitions (called ``atlases''), which yields robust and accurate results.


The proposed approach establishes a new benchmark model for autism classification on ABIDE, which is a particularly difficult dataset because of its heterogeneity, comprising subjects across a wide age range (5-64 years), and from sites that used different imaging protocols. Previous studies have reported cross-validated classification accuracies up to $67\%$ on ABIDE-I, the first phase of the ABIDE study~\cite{ABRAHAM2017}. 
The proposed CNN approach improves this accuracy to above $73\%$.
We also report, for the first time, independent test performance for benchmark and proposed models on the recently released second phase of ABIDE.

\section{Materials and methods}

\begin{figure}[htbp]
\hspace*{-0.5in}
  \centering 
  \includegraphics[width=5.5in]{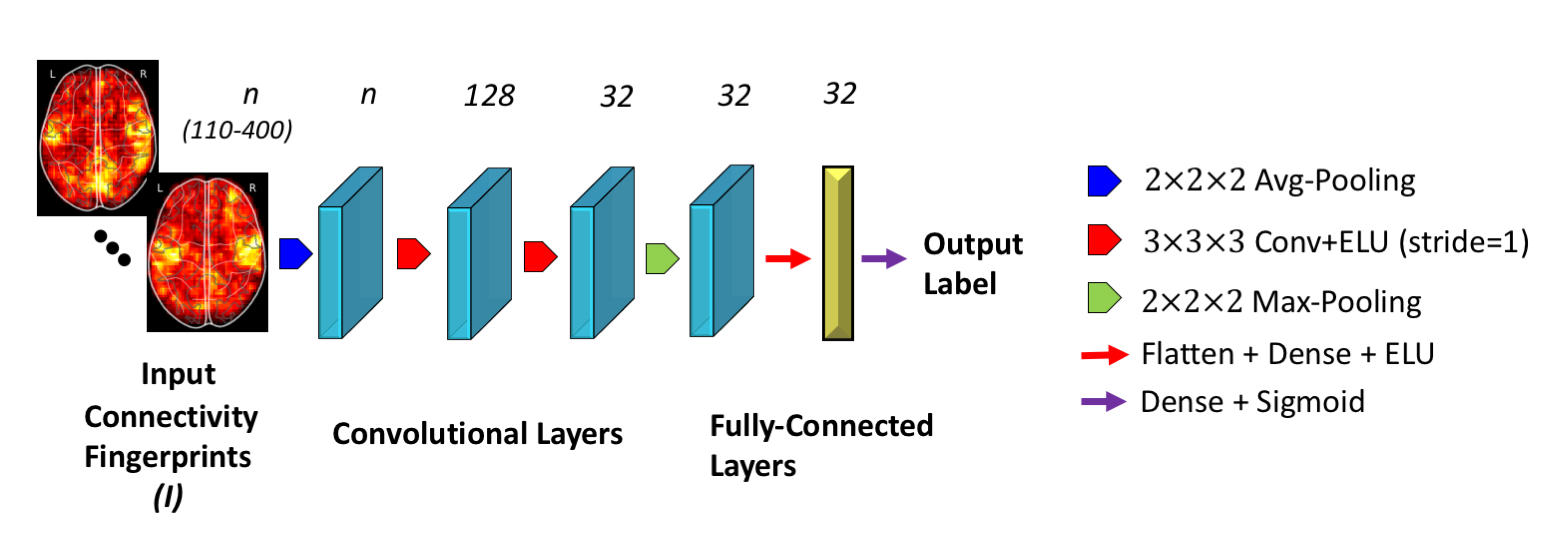} 
  \caption{Implemented CNN architecture. Number of channels denoted above.}
  \label{fig:nn} 
\end{figure}

\subsection{Proposed 3D CNN Approach}

Here, we present our strategy to adopt a CNN architecture for use with connectomic data.
The input to the CNN is formed by concatenating voxel-level maps of ``connectivity fingerprints'', which are represented as a multi-channel 3D volume. 
Each channel is a connectivity feature, such as the (Pearson) correlation between each voxel's time series and the average signal within a target ROI.
In our implementation, we use atlas-based brain parcellation schemes to define the target ROIs.
The total number of input channels thus represents the number of ROIs used for creating voxel-level fingerprints. 
We used a variety of so-called atlases, which define a specific parcellation of the brain into ROIs (see below for details).
Each atlas consisted of between 110 and 400 ROIs, where a larger number of regions corresponded to a finer scale parcellation.
For each atlas, we trained a separate model, which we report performance values for.
We also implemented an ensemble learning strategy, where the prediction was computed by taking a majority vote of the models corresponding to the different atlases.

In our experiments, we employed a simple CNN architecture, illustrated in Fig.~\ref{fig:nn}.
Our architecture has several convolutional layers, interspersed with max-pooling based down-sampling layers, followed by a couple of densely connected layers.
The models were trained for 50 epochs with a batch size of 64. The learning rate and momentum for Stochastic Gradient Descent (SGD) were set to 0.001 and 0.9 respectively. 
The same architecture and settings were used for all atlases. We note that each atlas is defined on a unique grey matter mask. To ensure that all classifiers (baseline and proposed) use information from the same voxels while computing mean ROI signals or connectivity patterns, respective gray matter masks of these atlas were used for masking the input image of connectivity fingerprints before feeding into the proposed convolutional architecture. 

\subsection{Baseline Methods}
Proposed CNN was compared against following baselines.
\vspace{-8pt}
\subsubsection*{Ridge Classifier:}
A linear regression model was trained with a loss function equal to the sum of squared differences between prediction and ground truth values and $\alpha$ times the squared norm of the weight vector.
The ground truth labels were encoded as $\pm$ 1 for the two output categories.
We test 10 linearly spaced values for the hyper-parameter $\alpha$ in the range [0.1,10] and report the highest cross-validation accuracy. 
Thus this baseline result reflects an \textit{optimistic} estimate of performance.
\vspace{-8pt}
\subsubsection*{Support Vector Machine:}
A linear SVM was trained to minimize $\beta$ times the squared hinge loss function plus the squared norm of the weight vector.
The hyper-parameter $\beta$ was tuned by maximizing cross-validation accuracy by searching over two orders of magnitude $([0.5,50])$.
As with the ridge classifier, this should be considered as an upper bound on generalization performance.
\vspace{-8pt}
\subsubsection*{Fully Connected Architecture:}
The fully-connected neural network (FCN) architecture takes as input functional connectivity estimates between pairs of ROIs, which is vectorized and processed bt a feed-forward network.
We implemented following architecture: 4 fully connected hidden layers, with 800, 500, 100 and 20 numbers of features and each linear layer followed by an elementwise Exponential Linear Unit (ELU) activation.
The output node was a sigmoid and computes disease probability, which is subsequently used for classification. 
The models were trained for 30 epochs with a batch size of 64. 
We monitored training curves to ensure that all trained models had converged before terminating the optimzation.
Stochastic Gradient Descent was used as the optimizer with learning rate and momentum set to 0.01 and 0.9 respectively. Dropout regularization parameter was set to 0.2 and applied to each layer during training.
\subsection{Experiments}
\subsubsection{Data:}
Autism Brain Imaging Data Exchange (ABIDE) is a multi-site open-access MRI study~\cite{DiMartino2017}. 
The first phase of ABIDE (ABIDE-I) compiled data from 1112 individuals, comprising 539 individuals diagnosed with Autism Spectrum Disorder (ASD) and 573 typical controls, from 17 sites. 
The second phase (ABIDE-II) was recently released, and consists of an additional 521 individuals with ASD and 593 healthy controls, from 19 sites. 

In our experiments, we used ABIDE-I subject data that passed manual quality assessments (QA) by all the functional raters. 
This yielded a final sample size of 774 ABIDE-I subjects, comprising 379 subjects with ASD and 395 typical controls. 
As an independent test dataset, we employed ABIDE-II subjects from sites that participated in ABIDE-I  and used the same MRI sequence parameters for data collection.
Since manual QA was not yet available for ABIDE-II, we performed an automatic quality control by selecting those subjects that retained at least 100 frames or 4 minutes of fMRI scans after motion scrubbing\cite{Power2014MethodsTD}. Motion scrubbing was performed based on Framewise Displacement (FD), discarding one volume before and two volumes after the frame with FD exceeding 0.5mm~\cite{Muschelli}. 
%
%
%
This step yielded a final ABIDE-II sample size of 163 individuals with ASD and 230 healthy controls. 

\subsubsection{Data Preprocessing:} 
ABIDE-I: The Preprocessed Connectomes Project (PCP) released preprocessed versions of ABIDE using several pipelines \cite{Cameron2013}. We used the data processed through Configurable Pipeline for Analysis of Connectomes (CPAC). This pipeline includes slice timing correction, motion correction, global mean intensity normalization and standardization of functional data to MNI space (3x3x3 mm resolution) before the extraction of ROI time series. Among the different versions of the release, data extracted with global signal regression and band-pass filtering (0.01-10Hz) was used in our analysis. 
%
%

ABIDE-II: We preprocessed the ABIDE-II dataset following the same sequence of steps listed for ABIDE-I in CPAC (v1.0.2a). 
Connectivity between distinct brain regions was estimated using Pearson's correlation coefficient. 



\subsubsection{Atlases:} In our experiments, we considered all atlases that were used for ROI time series extraction in PCP. 
These include the following seven atlases: Harvard-Oxford (HO), Craddock 200 (CC200), Eickhoff-Zilles (EZ), Talaraich and Tournox (TT), Dosenbach 160 (DOS160), Automated Anatomical Labelling (AAL) and Craddock 400 (CC400)~\cite{Dosenbach1358, DESIKAN2006968, AAL,CC,TT,EZ}.

For the baseline methods, each atlas was used to define a corresponding connectivity matrix which was fed as input to each model.
For the proposed model, the atlas ROIs were used as target ROIs to derive the input connectivity features at the voxel-level.
We also report results for an enesemble learning strategy, where we combined the predictions of models corresponding to individual atlases through majority voting to obtain improved and robust predictions. 


\subsubsection{Evaluation:}
We evaluated our model on the challenging task of autism classification using the two schemes. First, we implemented a 10-fold cross-validation scheme for ABIDE-I to be consistent with previously reported classification results~\cite{ABRAHAM2017,Plitt2015}.  
Second, we trained each model on all of ABIDE-I and computed test performance on an independent held-out set from ABIDE-II. This is used for assessing the generalization behavior of different classifiers. 
We report accuracy and the receiver operating curves (ROC), along with corresponding area under the curves (AUC) for each of these scenarios.

\begin{table}[tbp]
  \centering 
  
  \begin{tabular}{l|l|l|l|l|l} 
    Parcellation & Ridge Classifier & SVC (l2 penalty) & SVC (l1 penalty) & Deep Network & 3D-CNN \\ \hline
    HO  & 66.7/63.3 & 66.7/63.1 & 67.9/62.8 & 69.4/\textbf{67.7} &  \textbf{70.5}/\textbf{67.7} \\ \hline  
    CC200  & 69.7/67.4 & 69.5/68.7 & 68.8/66.4 & 70.5/71.5 & \textbf{71.2/72.8} \\ \hline 
    EZ  & 66.4/63.3 & 66.9/63.3 & 65.9/61.0 & 68.6/63.8 &  \textbf{69.3/66.4} \\ \hline 
    TT  & 64.4/66.1 & 65.3/66.7 & 64.3/61.3 & 67.1/65.9 &  \textbf{69.4/70.0} \\ \hline 
    CC400  & 70.2/69.4 & 70.5/69.7 & 67.5/68.1 & 71.0/69.9 & \textbf{71.7/70.5}  \\ \hline 
    AAL  & 65.4/63.3 & 65.7/62.3 & 68.1/62.6 & 66.7/65.4 &  \textbf{71.4/69.5} \\ \hline 
    DOS160  & 66.2/66.7 & 66.7/66.1 & 65.3/61.6 & 67.2/66.1 & \textbf{68.6/67.0} \\ \hline 
    Ensemble  & 69.8/66.7 & 69.6/67.1 & 70.1/64.2 & 71.5/69.9 & \textbf{73.3/71.7} \\ \hline \hline 
  \end{tabular}
  \newline 
  \caption{10-fold cross-validation on ABIDE-I/independent test on ABIDE-II accuracy of baseline models and proposed CNN approach. Best results are \textbf{bolded}.} 
  \label{tab:acc} 
\end{table}
 \begin{figure}[h] 
\centering 
\begin{minipage}{0.6\textwidth}  
  
  \includegraphics[width=\textwidth]{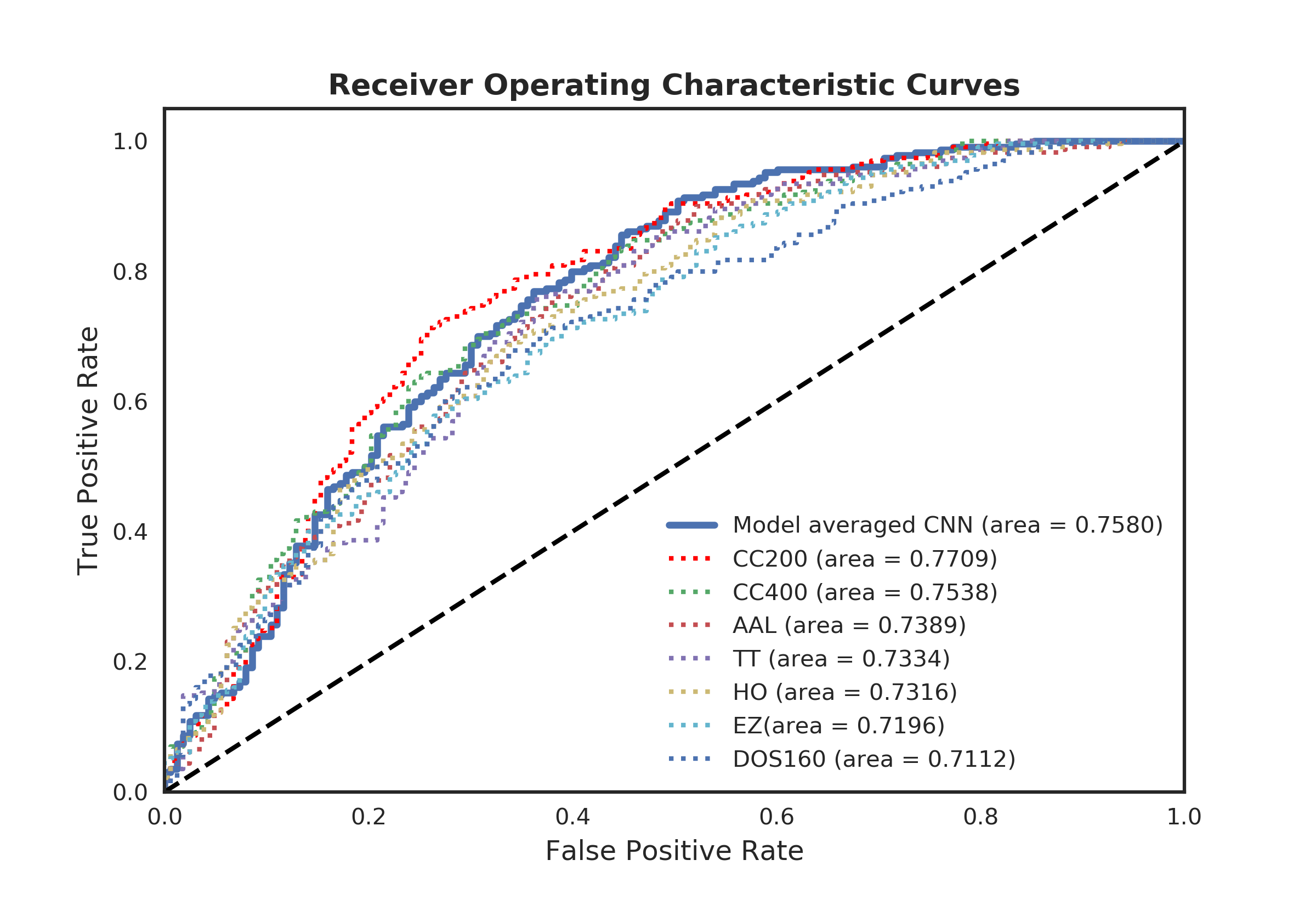} 
  \caption{ROC on independent ABIDE-II}
  \label{fig:results2auc} 
\end{minipage}
\end{figure} 
\section{Results}

Table \ref{tab:acc} shows cross-validation and independent test accuracy values for different models.
Proposed 3D CNN model consistently outperforms baselines. 
The ensemble CNN approach yields a classification accuracy of \textbf{73.3 \%} on ABIDE-I, significantly exceeding the state of the art~\cite{Heinsfeld18}.
Further, with an accuracy of \textbf{71.7 \%} on independent test data, the model also achieves good generalization.  
Figure~\ref{fig:results2auc} shows ROC curve obtained of individual atlases and their ensemble on ABIDE-II. The ensemble achieves an AUC of \textbf{75.8\%}.

\subsection{Visualization of CNN model}
\begin{figure}
\begin{subfigure}{.475\linewidth}
    \centering
    \includegraphics[width=\textwidth]{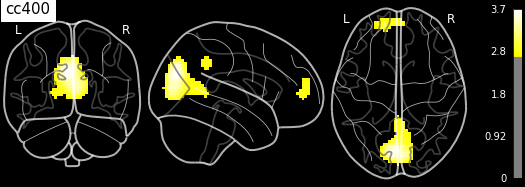} 
\end{subfigure}
\begin{subfigure}{.475\linewidth}
  \centering
  \includegraphics[width=\textwidth]{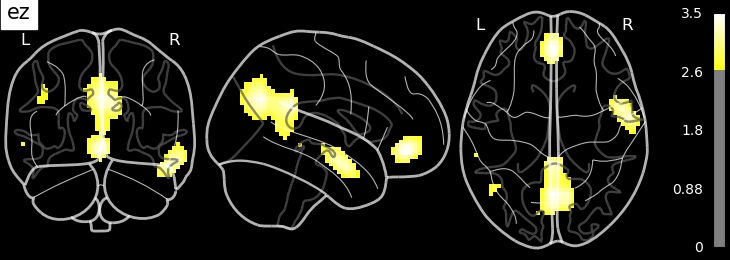} 
\end{subfigure} 

\begin{subfigure}{.475\linewidth}
    \centering
   \includegraphics[width=\textwidth]{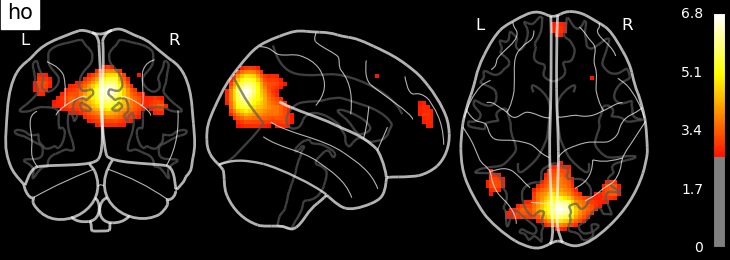} 
\end{subfigure}
\begin{subfigure}{.475\linewidth}
  \centering
  \includegraphics[width=\textwidth]{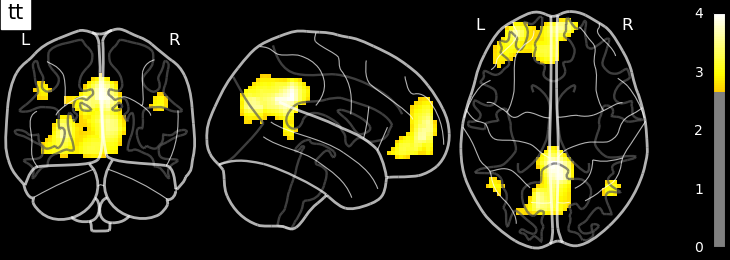} 
\end{subfigure} 

\begin{subfigure}{.475\linewidth}
    \centering
   \includegraphics[width=\textwidth]{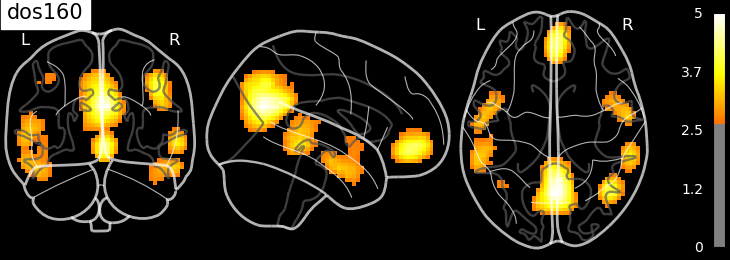} 
\end{subfigure}
\begin{subfigure}{.475\linewidth}
  \centering
  \includegraphics[width=\textwidth]{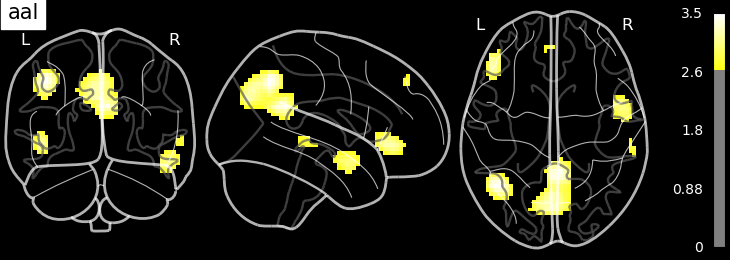} 
\end{subfigure} 
\hfill

\caption{CNN saliency maps averaged over ABIDE-II cases for different atlases.}
\label{fig:saliency}
\end{figure}

Visualization techniques for CNNs can help reveal salient features used by the model for discriminating between output classes. 
We employed the saliency map of~\cite{SimonyanVZ13}, which is a gradient-based technique. 
Essentially, this visualization approach computes the gradient of the output score with respect to the input image, i.e., the 3D volume, using a single backward pass through the trained neural network. 
We then computed voxel-level saliency as the maximum absolute gradient value across all input channels corresponding to different target ROIs. 
Fig~\ref{fig:saliency} shows these saliency maps averaged across all ABIDE-II cases for different atlases.

\section{Discussion}
In this paper, we presented a novel strategy to use 3D-CNN architectures for connectome classification. 
We conducted detailed empirical evaluations of the proposed model on a large dataset, which yielded significant improvements over state-of-the-art accuracy. 
In almost all cases, the performance of the proposed approach exceeded that of the baseline models, although the differences were modest for higher resolution parcellations. 

Another contribution of our paper is to highlight the advantage of ensemble learning, for example by majority voting over models corresponding to different atlases.
Atlases, or more generally ROIs, are often selected arbitrarily in the rs-fMRI community and our experiments demonstrate that averaging across these decisions can yield more robust and accurate predictions.

%

The interpretation of classification models is invaluable for biomedical applications, for example by offering biological insights or understanding the information that was used to make the prediction.  
Several previous studies have attempted to visualize abnormal connectivity patterns in disease. 
In this work, we present a strategy that allows us to interrogate the trained CNN models.
Our approach allows for visualizing the saliency map for a given individual, yet we leave the analysis of this for future work.
Instead we presented group-averaged maps for the different atlases.
As shown in Fig.~\ref{fig:saliency}, the saliency maps for the different atlases are rather consistent and highlight the so-called default mode network, which has been implicated in autism in prior studies~\cite{Padmanabhan2017}.
We also note some differences between the atlas saliency maps, which suggests that the different models are utilizing slightly different information content and thus can be complementary, explaining why model averaging can improve accuracy.




While the proposed CNN approach achieves promising accuracy on autism detection, there is room for further improvement.
We have not yet conducted a comprehensive optimization of the convolutional architecture.
Furthermore, there is likely more optimal choices than atlas-based target ROI correlations that are used as input to the model.
We envision an end-to-end learning strategy that can enable the optimization of these connectomic features.
 



\section{Conclusion}
Our experiments highlight the potential of deep neural network algorithms in the classification of functional connectomes and in expanding our understanding of brain disorders. When tailored for connectomes, modern DNN architectures like Convolutional Neural Networks offer an unparalleled opportunity to probe brain networks in disease.

\section{Acknowledgements}
This work was supported by NIH grants R01LM012719 (MS), R01AG053949 (MS), R21NS10463401 (AK), R01NS10264601A1 (AK), the NSF NeuroNex grant 1707312 (MS) and Anna-Maria and Stephen Kellen Foundation Junior Faculty Fellowship (AK).

%
%
\bibliography{connectome}

\begin{thebibliography}{10}

\bibitem{ABRAHAM2017}
Alexandre Abraham et~al.
\newblock Deriving reproducible biomarkers from multi-site resting-state data:
  An autism-based example.
\newblock {\em NeuroImage}, 147, 2017.

\bibitem{BrownH16}
Colin~J. Brown and Ghassan Hamarneh.
\newblock Machine learning on human connectome data from {MRI}.
\newblock {\em CoRR}, 1611.08699, 2016.

\bibitem{CC}
Craddock~R. Cameron et~al.
\newblock A whole brain fmri atlas generated via spatially constrained spectral
  clustering.
\newblock {\em Human Brain Mapping}, 33(8):1914--1928.

\bibitem{Cameron2013}
Craddock et~al.
\newblock {The Neuro Bureau Preprocessing Initiative: open sharing of
  preprocessed neuroimaging data and derivatives}.
\newblock {\em Frontiers in Neuroinformatics}, 2013.

\bibitem{DESIKAN2006968}
Desikan et~al.
\newblock An automated labeling system for subdividing the human cerebral
  cortex on mri scans into gyral based regions of interest.
\newblock {\em NeuroImage}, 31(3), 2006.

\bibitem{DiMartino2017}
Adriana Di~Martino et~al.
\newblock {Enhancing studies of the connectome in autism using the autism brain
  imaging data exchange II}.
\newblock {\em Scientific Data}, 4:170010, 2017.

\bibitem{Dosenbach1358}
Nico U.~F. Dosenbach, Binyam Nardos, Alexander~L. Cohen, et~al.
\newblock Prediction of individual brain maturity using fmri.
\newblock {\em Science}, 329(5997):1358--1361, 2010.

\bibitem{EZ}
Simon~B. Eickhoff et~al.
\newblock A new spm toolbox for combining probabilistic cytoarchitectonic maps
  and functional imaging data.
\newblock {\em NeuroImage}, 25(4), 2005.

\bibitem{Heinsfeld18}
Anibal~S{\'o}lon Heinsfeld et~al.
\newblock Identification of autism spectrum disorder using deep learning and
  the abide dataset.
\newblock In {\em NeuroImage: Clinical}, 2018.

\bibitem{Kaiser2011}
M.~{Kaiser}.
\newblock {A Tutorial in Connectome Analysis: Topological and Spatial Features
  of Brain Networks}.
\newblock {\em ArXiv e-prints}, May 2011.

\bibitem{Kawahara2017}
J.~Kawahara et~al.
\newblock {BrainNetCNN: Convolutional neural networks for brain networks;
  towards predicting neurodevelopment}.
\newblock {\em NeuroImage}, 146, Feb 2017.

\bibitem{kong2018spatial}
Ru~Kong et~al.
\newblock Spatial topography of individual-specific cortical networks predicts
  human cognition, personality, and emotion.
\newblock {\em Cerebral Cortex}, 2018.

\bibitem{TT}
Lancaster~Jack L., Woldorff~Marty G., et~al.
\newblock Automated talairach atlas labels for functional brain mapping.
\newblock {\em Human Brain Mapping}, 10(3):120--131.

\bibitem{Mennes2012}
Mennes et~al.
\newblock {Resting State Functional Connectivity Correlates of Inhibitory
  Control in Children with ADHD}.
\newblock {\em Front Psychiatry}, 2012.

\bibitem{Muschelli}
John Muschelli, Mary~Beth Nebel, et~al.
\newblock Reduction of motion-related artifacts in resting state fmri using
  acompcor.
\newblock {\em NeuroImage}, 96, 2014.

\bibitem{pmlr-v48-niepert16}
Mathias Niepert et~al.
\newblock Learning convolutional neural networks for graphs.
\newblock Proceedings of Machine Learning Research, New York, USA, Jun 2016.

\bibitem{Padmanabhan2017}
Aarthi Padmanabhan et~al.
\newblock {The Default Mode Network in Autism}.
\newblock {\em Biological Psychiatry: Cognitive Neuroscience and Neuroimaging},
  2(6), 2017.

\bibitem{Plitt2015}
Plitt et~al.
\newblock {Functional connectivity classification of autism identifies highly
  predictive brain features but falls short of biomarker standards}.
\newblock {\em NeuroImage:Clinical}, 2015.

\bibitem{Power2014MethodsTD}
Jonathan~D. Power, Anish Mitra, et~al.
\newblock Methods to detect, characterize, and remove motion artifact in
  resting state fmri.
\newblock {\em NeuroImage}, 84, 2014.

\bibitem{SimonyanVZ13}
Karen Simonyan et~al.
\newblock Deep inside convolutional networks: Visualising image classification
  models and saliency maps.
\newblock {\em CoRR}, 1312.6034, 2013.

\bibitem{AAL}
N.~Tzourio-Mazoyer et~al.
\newblock Automated anatomical labeling of activations in spm using a
  macroscopic anatomical parcellation of the mni mri single-subject brain.
\newblock {\em NeuroImage}, 15(1), 2002.

\bibitem{varoquaux10}
Ga{\"e}l Varoquaux et~al.
\newblock Detection of brain functional-connectivity difference in post-stroke
  patients using group-level covariance modeling.
\newblock In {\em MICCAI 2010}.

\end{thebibliography}
\bibliographystyle{plain}
%


%
%

\clearpage


\end{document}